\documentclass[journal]{IEEEtran}
\usepackage{cite}
\usepackage{amsmath,amssymb,amsfonts}
\usepackage{graphicx}
\usepackage{textcomp}
\usepackage{multirow}
\usepackage{algorithm2e}
\usepackage{algpseudocode}
\usepackage{hyperref}
\usepackage{xcolor}
\newtheorem{remark}{Remark}
\ifCLASSINFOpdf
\else
\fi
\begin{document}
%
% paper title
% Titles are generally capitalized except for words such as a, an, and, as,
% at, but, by, for, in, nor, of, on, or, the, to and up, which are usually
% not capitalized unless they are the first or last word of the title.
% Linebreaks \\ can be used within to get better formatting as desired.
% Do not put math or special symbols in the title.
\title{MATNilm: Multi-appliance-task Non-intrusive Load Monitoring with Limited Labeled Data}
%
%
% author names and IEEE memberships
% note positions of commas and nonbreaking spaces ( ~ ) LaTeX will not break
% a structure at a ~ so this keeps an author's name from being broken across
% two lines.
% use \thanks{} to gain access to the first footnote area
% a separate \thanks must be used for each paragraph as LaTeX2e's \thanks
% was not built to handle multiple paragraphs
%

\author{Jing Xiong,
        Tianqi Hong,
        Dongbo Zhao,
        and Yu Zhang% <-this % stops a space

\thanks{Jing Xiong is with the University of California Santa Cruz, Santa Cruz, CA 95064 USA (e-mail: jxiong20@outlook.com). }
\thanks{Tianqi Hong is with the School of Electrical and Computer Engineering at the University of Georgia, Athens, GA 30602, USA (e-mails: Tianqi.Hong@uga.edu).}
\thanks{Dongbo Zhao is with Eaton Research Labs, Golden, CO 80401 USA (e-mail: zhaodongbo2008@gmail.com).}
\thanks{Yu Zhang is with the University of California Santa Cruz, Santa Cruz, CA 95064 USA (e-mail: yzhan419@ucsc.edu). }}

\maketitle

\begin{abstract}
Non-intrusive load monitoring (NILM)  identifies the status and power consumption of various household appliances by disaggregating the total power usage signal of an entire house. Efficient and accurate load monitoring facilitates user profile establishment, intelligent household energy management, and peak load shifting. This is beneficial for both the end-users and utilities by improving the overall efficiency of a power distribution network. Existing approaches mainly focus on developing an individual model for each appliance. Those approaches typically rely on a large amount of household-labeled data which is hard to collect. In this paper, we propose a multi-appliance-task framework with a training-efficient sample augmentation (SA) scheme that boosts the disaggregation performance with limited labeled data. For each appliance, we develop a shared-hierarchical split structure for its regression and classification tasks. In addition, we also propose a two-dimensional attention mechanism in order to capture spatio-temporal correlations among all appliances. 
With only one-day training data and limited appliance operation profiles, the proposed SA algorithm can achieve comparable test performance to the case of training with the full dataset. Finally, simulation results show that our proposed approach features a significantly improved performance over many baseline models. The relative errors can be reduced by more than 50\% on average. The codes of this work are available at \url{https://github.com/jxiong22/MATNilm}.
\end{abstract}


\section{Introduction}

\IEEEPARstart{E}{xploring} users' power usage patterns and providing energy saving guidance is important for smart power grids. Users can participate in power production activities through renewable energy generation and effectively improve power efficiency by adjusting power usage patterns to shift the peak load demand \cite{hosseini2017non}. It's relatively easy to install a meter that monitors the household power consumption every few seconds, or at a higher sampling rate. However, recording the power usage of each appliance can be inconvenient and costly. In this context, non-intrusive load monitoring (NILM), also known as energy disaggregation, aims to leverage a household's aggregate power consumption signal to make inferences about individual appliance usage without using extra sensors. Based on the measurement sampling rate, NILM can be divided into two categories. High-frequency measurements are in the order of 10 kHz, including various features such as harmonics \cite{lin2013development} and V-I trajectory \cite{liu2018non, wang2018non}.  This type of data requires sophisticated hardware and suffers from a higher cost. NILM via low-frequency data is a more challenging task. As the sampling rate drops below 1 Hz, distinct features used in the high-frequency data are no longer available. Therefore, extracting high-level hidden features is the key to achieving better performance. In this work, we focus on low-frequency data and propose a solution when labeled data are limited.

\subsection{Prior Work}

NILM was first studied by Hart in 1992 \cite{hart1992nonintrusive}. Afterward, many classical models were developed to deal with the problem. \cite{ghahramani1997factorial} proposes the factorial hidden Markov model (FHMM). Based on that work, researchers design a conditional factorial hidden semi-Markov model that can incorporate extra features about the appliance status in \cite{kim2011unsupervised}. More classical models are summarized in review papers \cite{bonfigli2015unsupervised, zhuang2018overview, najafi2018data}.

Over the past few years, deep neural networks (DNNs) have demonstrated remarkable achievements in various fields, such as natural language processing and computer vision. Several studies in NILM have also delved into the utilization of DNNs, as highlighted in the work by Bousbiat et al. \cite{bousbiat2022unlocking}. Classic DNN layers such as convolutional neural network (CNN) \cite{moradzadeh2021practical}, recurrent neural network (RNN) \cite{zhang2018sequence} and temporal convolutional network (TCN) \cite{liu2021samnet} have been investigated extensively. In \cite{kaselimi2020context}, the authors propose a bidirectional long short-term memory (LSTM) model to capture the context information of the aggregated consumption. A Bayesian-optimized framework is utilized to select the best configuration of the proposed regression model. In \cite{feng2022multi},  a parallel CNN and a bidirectional gated recurrent unit network are developed to extract spatial-temporal features whose weights are updated through an attention mechanism. In \cite{tanoni2022multilabel}, the NILM task is modeled as a multiple-instance learning problem, and a convolutional recurrent neural network is trained with weak supervision. 

An increasing number of works have investigated advanced DNNs for NILM.
In \cite{shin2019subtask}, the authors develop a subtask gated network (SGN). The regression subnetwork outputs the electricity usage while the classification counterpart, which is used as a gating unit, specifies the on/off state for each given appliance. Based on this work, \cite{piccialli2021improving} enhances the model's effectiveness by incorporating an attention mechanism into the regression subnetwork. This integration intensifies the network's ability to identify and focus on the most pertinent parts of the input sequence, thereby amplifying its representational prowess. \cite{chen_2020} develops an architecture using different receptive field size branches. Branch-wise gates connect corresponding branches in two subnetworks. To improve the performance, the authors employ a Wasserstein generative adversarial network (WGAN) with gradient penalty during the model's training. The other GAN structures are also explored. \cite{kaselimi2020energan++} proposes an EnerGAN++ structure for the NILM task. 
An autoencoder is implemented as the GAN generator, enabling a nonlinear signal source separation. The aggregate input concatenating with actual or estimated appliance consumption is the discriminator's input. In \cite{liu2021unsupervised}, the adversarial training process and the joint adaptation loss are introduced to leverage the hidden information from unlabeled data.

\subsection{Single-appliance vis-\`{a}-vis Multi-appliance}

There are different setups for NILM, which can be categorized based on the learning tasks. Typically, the regression task involves the inference of individual appliance's power usage. The classification task aims to determine the appliance's on/off status or different states related to power levels \cite{harell2019wavenilm}. As summarized in the previous section, various machine learning approaches have been proposed for both tasks. In addition, tools such as pattern matching \cite{liu2017dynamic} and source separation \cite{rahimpour2017non} can also be found in the literature; see details in \cite{schirmer2022non}. 

Given an aggregate power consumption signal, a learning model can be trained for each appliance separately, or it can be used to estimate the power signals for all appliances in consideration. We use the terms ``single-appliance" and ``multi-appliance" to refer to those two approaches, respectively. 
Most of the existing regression models adopt the single-appliance approach, which ignores the interactions among different appliances. For example, it is uncommon to operate a dishwasher and a microwave at the same time. The usage of a dryer often comes after a washer. In contrast, such context-aware information can be implicitly incorporated by the multi-appliance approach into the modeling process, which can improve the disaggregation accuracy.

The multi-appliance approach of previous works focuses on the  on/off status and state classification. For instance,  a semi-supervised multi-label deep learning framework is proposed in \cite{yang2019semisupervised} to monitor the on/off status of multiple appliances simultaneously. In \cite{mueller2016accurate}, an HMM-based model is designed to infer appliance states over time by eliminating the possibility of unmodeled loads and outputting the used energy. A WaveNet, which is fed with active, reactive and apparent power as inputs, is used to estimate appliance states and currents \cite{harell2019wavenilm}. In addition, an effort has been put forth to estimate the total energy consumption of a set of appliances with similar operation patterns \cite{liu2016non}.

The direct regression task in multi-appliance setups is still an ongoing development, with research studies such as \cite{shin2019subtask} and \cite{kelly2015neural} highlighting its advantages over state classification. Direct regression allows for more accurate and detailed estimations for individual appliances. However, existing approaches have certain limitations, which include the vanilla neural network for multi-appliance (VMA) in \cite{schirmer2021binary}, the U-Net architecture in \cite{faustine2020unet}, and the transfer learning techniques employed in \cite{li2023transfer}. These methods directly output power consumption for multiple appliances from the output layer without explicitly considering the inter-appliance relationships at different time steps. This limitation hinders their effectiveness in achieving more accurate estimations in real-world scenarios.

\subsection{Data Augmentation}

Data augmentation for improving NILM performance has been explored in the literature. \cite{kong2019practical} generates synthetic training data by randomly adding the appliance profile to the user's aggregate load. Besides the on-duration period of appliances, \cite{rafiq2021generalizability} also considers the off-duration when generating augmented synthetic data. However, it is hard to determine how much synthetic data is really needed. On the other hand, this type of data augmentation (DA) only incorporates existing appliance profiles, which are very limited compared with the potentially vast models of each appliance. In order to generate more appliance profiles, \cite{chen_2020} proposes an on-state augmentation method by adding extra noise to the fridge and the corresponding aggregation input. Existing data augmentation algorithms are mainly designed for generating a synthetic dataset for each aggregate and appliance pair, which acts as a regularizer for each appliance separately, for increasing the generalization capability of the model \cite{kelly2015neural}. Moreover, these methods are still trained by a full dataset, which may not fit the limited data scenario. 

Some existing works also analyze the limited data scenario in the testing phase. 
In \cite{cimen2022deep}, the author compares the energy disaggregation performance with 100\%, 50\%, and 25\% of each house data in the REDD dataset. In \cite{chen_2020}, the author shows the dishwasher and microwave performance of the proposed network with 20\% of the training data and fridge with 5\% and 10\% of the training data. These works do not focus on data-limited scenarios, but validate the effects of data-limited scenarios on the proposed model. Therefore, there is a lack of sufficient mining of the conditions of limited data scenarios. In \cite{li2018residential}, a semi-supervised graph-based approach is proposed with a limited amount of ground-truth data and a large pool of unlabeled observations to classify the activation status of the appliances accurately. A novel LSTM combined with a probabilistic neural network (PNN) algorithm is proposed to classify the active appliance type in \cite{zhou2020novel}. The authors leverage the transfer learning techniques to lower the requirements for training data further. However, these works only focus on the multi-label classification task. The regression task with limited labeled data still lacks practical solutions.

\subsection{Contribution}

In this paper, we address the challenges of performing NILM with limited data. We test extreme cases by using only one-day training data and limited operation profiles. The main novelty and contributions of our work are summarized as follows:
\begin{itemize}
\item This study presents an effective framework that leverages one-day training data to estimate both power consumption (regression) and appliance status (classification). The numerical tests demonstrate competitive performance of our proposed approach when it is compared with state-of-the-art models trained on the full datasets.
% \item To overcome the challenges posed by limited data scenarios, we introduce a sample augmentation algorithm that incorporates data augmentation directly into the training process. This approach generates diverse training samples until no further performance improvements are observed, eliminating the need for traditional synthetic dataset generation prior to training.
\item To overcome the challenges posed by limited data, we introduce a novel on-the-fly sample augmentation scheme that does not pre-select the amount of needed data samples. This approach enables end-to-end learning by integrating data augmentation into the training process. Compared with existing data augmentation in NILM, our algorithm is able to efficiently generate diverse training samples until no further performance improvements.
\item To enhance the performance of NILM, we propose a novel multi-appliance-task (MAT) architecture with a two-dimensional multi-head self-attention (2DMA) mechanism. Our design captures power consumption relationships both horizontally across multiple appliances and vertically across time steps. Our approach yields significant improvements over conventional single-appliance approaches, which do not account for unique characteristics of the NILM problem.
\end{itemize}

The remainder of the paper is organized as follows: Section II is the problem formulation and notations. Section III presents the detailed algorithm for sample augmentation and MAT structure. Simulation setup and results are reported in Sections IV and V, respectively. Finally, Section VI summarizes this work.
% The source code of the proposed framework is available at \href{https://github.com/xiongjingzhuhai/MAT.git}{https://github.com/xiongjingzhuhai/MAT.git}.

\section{Preliminary}

Machine learning models for NILM aim to learn a mapping from aggregated power consumption (the input signal) to individual appliances' power consumption (the output signal). In order to train a model, a household labeled set of input (aggregate power consumption) - output (appliances power consumption) pairs $\left\{\left(x_i, \mathbf{y}_i\right)\right\}_{i=1}^{T_l}$ are used where $T_l$ is the number of time steps.
Typically, a sliding window method is used to construct the training dataset $$\mathcal{D}=\left\{\left(\mathbf{x}_{t:t+T-1}, \mathbf{y}_{t+w:t+T-w-1}\right)\, |\, t \in \left\{s*k\right\}_{k=1}^{N} \right\},$$ 
where $T$ is the input length, $s$ is step size, $N$ is the number of training examples and $w$ is the additional window size. For context-aware setup \cite{kaselimi2020context}\cite{shin2019subtask}\cite{chen_2020}, we have $w > 0$, which means the length of the input is longer than the length of the output, providing 'extra' input context to output. 

% Denote $\mathcal{A}$ dd $\mathcal{K}$ as the appliance set to be disaggregated. 
Let $\mathbf{x}_{t:t+T-1}=(x_t,x_{t+1},\dots,x_{t+T-1})^{\top} \in \mathbb{R}^{T}$ denotes the aggregate household load consumption, $\mathbf{y}_{t:t+T-1}^i=(y^i_t,y^i_{t+1},\dots,y^i_{t+T-1})^{\top} \in \mathbb{R}^T$ denotes the corresponding power consumption of appliance $i$ to be disaggregated in the same period, the remaining unlabeled appliance consumption as $\mathbf{u}_{t:t+T-1}=(u_t,u_{t+1},\dots,u_{t+T-1})^{\top} \in \mathbb{R}^{T}$. Thus, the aggregate load consumption at each time step $t$ would be the summation of appliance load consumption and unknown appliance load consumption:
\begin{equation} \label{e1}
x_t = \sum_{i \in \mathcal{K}} y_t^i + u_t + \epsilon_t,
\end{equation}
where $\epsilon$ is the measurement noise, $\mathcal{K}$ is the set of appliances to be disaggregated.

For the NILM task, given the input as the aggregate load consumption $\mathbf{x}$, we aim to separate the signal to each appliance $\mathbf{y}^1, \mathbf{y}^2, \dots, \mathbf{y}^{|\mathcal{K}|}$. The majority of existing methods develop $|\mathcal{K}|$ independent models to learn each appliance's mapping:
\begin{equation} \label{e2}
\hat{\mathbf{y}}_{t+w:t+T-w-1}^i = f^i(\mathbf{x}_{t:t+T-1}).
\end{equation}

We define a multi-appliance-task training schema
\begin{multline} \label{e3}
\{\hat{\mathbf{y}}_{t+w:t+T-w-1}^1, \hat{\mathbf{y}}_{t+w:t+T-w-1}^2, \dots, \hat{\mathbf{y}}_{t+w:t+T-w-1}^{|\mathcal{K}|}\}\\
= f(\mathbf{x}_{t:t+T-1}),
\end{multline}
where $f$ is trained jointly by all appliances in set $\mathcal{K}$. 

\section{Methodology}
In this section, we present our proposed framework for NILM with limited household labeled data. We first introduce a novel sample augmentation scheme, which is followed by the MAT structure with 2DMA.

\subsection{Sample Augmentation (SA) Algorithm}

A deep learning model typically needs a massive amount of training data to achieve good generalization capability. ``Massiveness" means not only the number of training samples but also data diversity. Existing approaches rely on numerous household labeled data $\left\{\left(x_i, \mathbf{y}_i\right)\right\}_{i=1}^{T_l}$. However, collecting labeled data is not easy and often requires the professional installation of hardware. It can be time-consuming and brings up privacy concerns.  In addition, the collected dataset may have all kinds of issues such as time inconsistency as well as missing and bad data. As some appliances remain OFF most of the time, the dataset can be highly imbalanced. Finally, the labeled samples are often limited and not rich enough compared with a variety of appliance models and usage patterns.  

To address the aforementioned challenges, we propose a dynamic sample augmentation algorithm to generate appliance profiles based on real data. As shown in \eqref{e1}, aggregate data can be synthetically generated by the combination of the appliance's operation profiles. An appliance's operation profile is defined as a sequence of sampled power measurements over one complete operation cycle \cite{kelly2015neural}. Based on this observation, we can generate synthetic labeled data $\left\{\left(x_i, \mathbf{y}_i\right)\right\}_{i=1}^{T_l}$ according to comprehensive operation profiles, which are assumed to be available. The proposed algorithm augments the training sample adaptively until new instances cannot improve the performance further.

Our algorithm is based on a newly constructed appliance pool $\mathcal{A} = \cup_{i=1}^{N_a}\mathcal{A}_i$, which collects power consumption signals of $N_a$ appliances; see also \cite{kong2019practical}.
For appliance $i$, $\mathcal{A}_i$ is the set $N_i$ operation profiles $\{\mathbf{\tilde{y}}^i_j\}_{j=1}^{N_i}$. The appliance pool can be built by  i) real operation profiles where measurements are provided by existing datasets, appliance companies, or smart plugs; and
ii) synthetic operation profiles produced by generative models such as GAN, diffusion models, or other algorithms.

To further diversify the training data, we utilize time series augmentation methods to modify existing profiles via vertical scale, horizontal scale, and mixed vertical-horizontal scale. 
\begin{itemize}
    \item Vertical scaling: the signal magnitude of $\mathbf{y}^i$ is scaled by $\alpha$ times, i.e., $\mathbf{y}^i \times \alpha$, where $\alpha \sim N(1, \sigma^{2})$ is a Gaussian random variable with mean one and variance $\sigma^{2}$. 
    \item Horizontal scaling: the signal length $l$ of $\mathbf{y}^i$ is scaled (compressed or extended) by $\beta$ times, i.e., $l \times \beta$, where $\beta \sim N(1, \sigma^{2})$. 
    \item Linear interpolation is used to generate unknown data points while scaling. 
\end{itemize}
Other modification methods could be added accordingly.

Let $\text{prob}^i$ denote the probability that the profile of appliance $i$ gets sample augmentation. If needed, we can tune this parameter to balance the on/off sampling rate. 
In addition, define $p^i = [p^i_\mathrm{IN},p^i_\mathrm{V},p^i_\mathrm{H},p^i_\mathrm{M}]$ as the probabilities of four modes: intact, vertical scaling, horizontal scaling and mixed scaling.
The SA algorithm is applied once a batch of samples is selected during training. As the training continues, new training samples will be continuously generated, which may improve the generalization capability of a machine learning model. The training will stop once the new training instance cannot contribute to the performance improvement, or the maximum training epoch has been reached. The detailed implementation of the proposed sample augmentation is given in Algorithm~\ref{alg:dataAugAl}. 
% Algorithm \ref{alg:getSignal} illustrates the procedure to get a sliced signal in the appliance profile pool.

\begin{remark}[Merits of sample augmentation (SA)]
The proposed SA offers several advantages over existing data augmentation methods that generate a synthetic dataset. First, it does not require a predefined size of the augmented dataset. SA with early stopping criteria enables an efficient generation of diverse training data until no further performance improvements. This property is not present in existing augmentation methods that often rely on ad-hoc or trial-and-error ways to determine the amount of needed data samples. Second, performing SA within the training process enables end-to-end training of the model. This simplifies the training pipeline and eliminates the need for a separate data preprocessing step. It makes the training more efficient with reduced memory overhead. Compared with existing DA schemes that require generating and storing a separate augmented dataset, end-to-end SA generates augmented data on the fly during the training. This further enables the model to learn from a diverse range of augmented data and avoid overfitting. Finally, in our scenario of limited data, existing DA approaches need to generate a synthetic dataset by creating multiple copies of the original dataset, based on which augmentation techniques can be applied. This approach is computationally expensive and time-consuming. In contrast, the proposed SA does not require an increase in the length of the dataset, which is more computationally efficient.
\end{remark}

\RestyleAlgo{ruled}
% \SetKwComment{Comment}{/* }{ */}

\begin{algorithm}[bt!]
\caption{Sample Augmentation Algorithm}
\label{alg:dataAugAl}
\SetKwInOut{Input}{Input}
\SetKwInOut{Output}{Output}
\Input{Original mini-batch $\Bar{\mathcal{D}}$ from training set $\mathcal{D}$; \newline 
Pre-defined sample augmentation probability for each appliance $\text{prob}^i$; \newline
Pre-defined modify-mode probability mass function for each appliance $p^i$; and \newline
Appliance pool $\mathcal{A}$
}
\Output{Augmented mini-batch $\Bar{\mathcal{D}}_a$}

\For{$(\mathbf{x},\mathbf{y},\mathbf{y}_c) \in \Bar{\mathcal{D}}$}{
  \For{appliance $i$ in $\mathcal{A}$}{
        $r \gets$ generate a random number from a uniform distribution over $[0,1]$\;
      \If{$r < \mathrm{prob}^i$}{
        $sig \gets$ randomly select an operation profile $\mathbf{\tilde{y}}^i_j$ from $\mathcal{A}_i$\;
        $mode \gets$ randomly choose a modification mode based on the distribution $p^i$\;
        $sig\_new \gets$ modify $sig$ according to the chosen mode, and randomly slice $sig\_new$ to the same length of $\mathbf{y}^i_j$\;
        % ************************** select sig new portion *******************
        
        \If{$i \in \mathcal{K}$}{$\mathbf{X} \gets \mathbf{X} - \mathbf{y}^i_j$\;
        % $\mathbf{y}^i \gets \mathbf{0}$\;
        $\mathbf{y}^i \gets sig\_new$\;      
        $\mathbf{y}^i_c \gets $ update on/off status\;}
        $\mathbf{X} \gets \mathbf{X} + sig\_new$\;
   
}
}
}
\end{algorithm}

\subsection{Multi-appliance-task Network Architecture}
\begin{figure}[!tb]
\centering
\includegraphics[width=0.9\columnwidth]{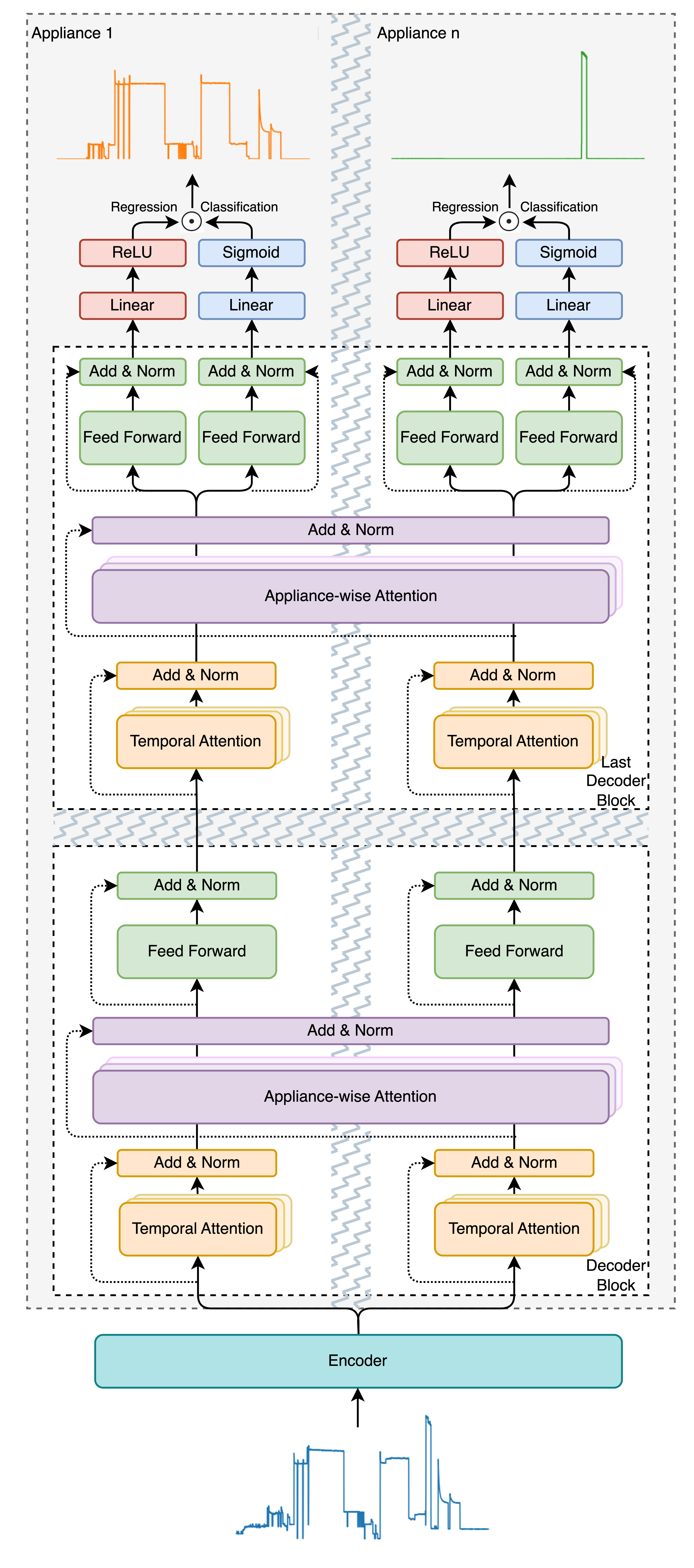} % Reduce the figure size so that it is slightly narrower than the column. Don't use precise values for figure width.This setup will avoid overfull boxes.
\caption{Multi-appliance-task network architecture with temporal attention and appliance-wise attention. The decoder comprises a stack of $m$ identical decoder blocks with $n$ branches. $m$ and $n$ are the numbers of decoder blocks and the number of appliances to be disaggregated, respectively. This figure only shows the case when $n=2$ and $m=1$. The grey zigzag areas can be extended as $n$ and $m$ increase.}
\label{fig:overall_structure}
\end{figure}

Multi-task learning is able to enhance the model generalization capability and performance when the tasks share similar knowledge and could be leveraged by one and the other. In the NILM task, the model is acted as a resource allocator that distributes the aggregate power into each appliance. From this point of view, the power 'assigned' to one appliance is dependent on the power 'assigned' to the others, since the summation of each disaggregated value cannot exceed the aggregate input inherently. In order to model this connection, we design a MAT network with 2DMA. We use an encoder and decoder structure as its backbone where the encoder learns a shared representation for all appliances and the decoder tries to distribute the power into each appliance. The decoder is split into $n$ branches, and each branch is further split into two branches for regression and classification tasks. $m$ decoder block is stacked in the decoder to extract higher-level features from the shared representation. And each decoder block is composed of a 2DMA layer and a fully connected feed-forward layer. Following the design from the transformer, we also employ residual connection and layer normalization for each sub-layer. The overall structure is shown in Fig. \ref{fig:overall_structure}. 

\subsubsection{2DMA mechanism}
The 2DMA mechanism comprises two attention layers: temporal attention and appliance-wise attention. The temporal attention layer is specifically designed to capture the temporal correlations across time for a given appliance by selectively attending to different time steps within the input sequence. This layer enables the model to effectively capture temporal dependencies. On the other hand, the appliance-wise attention layer is implemented to exploit the representations of other appliances at the same time step.
\\
\indent The 2DMA architecture is built based on the multi-head (MH) attention, which is a powerful mechanism for exploring complex relationships among input elements. 
Specifically, multi-head attention is defined as \cite{vaswani2017attention}:
\begin{subequations} \label{eq:MA}
\begin{align}
&\operatorname{Attention}(\mathbf{Q},\mathbf{K},\mathbf{V})=\operatorname{Softmax}\left(\frac{\mathbf{Q}\mathbf{K}^{\top}}{\sqrt{d_{k}}}\right) \mathbf{V},\label{eq:SDPatten} \\
&\operatorname{head}_{\mathrm{i}}=\operatorname{Attention}\left(\mathbf{Q}\mathbf{W}_{i}^{Q},\mathbf{K}\mathbf{W}_{i}^{K},\mathbf{V}\mathbf{W}_{i}^{V}\right),\label{eq:head}\\
&\operatorname{MH}(\mathbf{Q},\mathbf{K},\mathbf{V})=\operatorname{Concat}\left(\operatorname{head}_{1},\ldots, \operatorname{head}_{\mathrm{h}}\right) \mathbf{W}^{O}. \label{eq:headComb}
\end{align}
\end{subequations}

Equation \eqref{eq:SDPatten} is the scaled dot-product attention that takes the input in the form of three matrices, known as the query $\mathbf{Q}$, the key $\mathbf{K}$, and the value $\mathbf{V}$. 
Essentially, the attention function maps each query and a set of key-value pairs to an output, which is computed as a weighted sum of the values. The weight assigned to each value is given by a similarity score between the query and the corresponding key.
Then, we perform the attention computation in parallel with different projected versions of queries, keys and values to produce multiple attention heads in \eqref{eq:head}.  
To this end, those independent attention heads are concatenated and transformed with another learned linear projection $\mathbf{W}^{O}$ to obtain the final output \eqref{eq:headComb}. 
Such a linear projection, which serves to reduce the dimension of the long concatenated head, learns which attention heads to take more notice of. 
All above matrices $\mathbf{W}$ are learnable parameters obtained by e.g., backpropagation.

The self-attention learns how best to route information between pieces (known as tokens) of an input sequence while the multi-head version enables the model to jointly attend to information from different representation subspaces at different positions \cite{vaswani2017attention}.
This ubiquitous mechanism facilitates the creation of richer and better representations that can boost performance on various machine learning tasks.
\\
\indent In the temporal attention, $\mathbf{Q}, \mathbf{K}, \mathbf{V}$ come from the hidden representations of the same appliance learned from the previous layer $\mathbf{h}_{t+w:t+T-w-1}^i$ and are updated across different time steps:
\begin{multline}
\boldsymbol{\theta}_{t+w:t+T-w-1}^i
\\=\operatorname{LayerNorm}\left(\mathbf{h}_{t+w:t+T-w-1}^i + \operatorname{MH}\left(\mathbf{h}_{t+w:t+T-w-1}^i\right)\right).
\end{multline}
\\
\indent Similarly, in the appliance-wise attention, $\mathbf{Q}, \mathbf{K}, \mathbf{V}$ come from the hidden representations of each appliance at the same time step $\boldsymbol{\theta}_t^{i \in \mathcal{K}}$. Each appliance can update its representation based on other appliances' representations:
\begin{equation}
\boldsymbol{\phi}_{t}^{i \in \mathcal{K}}=\operatorname{LayerNorm}\left(\boldsymbol{\theta}_t^{i \in \mathcal{K}} + \operatorname{MH}\left(\boldsymbol{\theta}_t^{i \in \mathcal{K}}\right)\right).
\end{equation}

\subsubsection{MATNilm architecture}
Given an aggregated power consumption signal $\mathbf{x}_{t:t+T-1}$, the encoder learns a hidden shared representation for better separating by later appliance-specific split task
\begin{equation}
\mathbf{h}_{t+w:t+T-w-1}=\operatorname{Encoder}\left(\mathbf{x}_{t:t+T-1}\right),
\end{equation}
where $w=0$ is for general setup and $w>0$ is for context-aware setup.
The decoder splits a branch for each appliance $i \in \mathcal{K}$ with initial input $\mathbf{h}_{t+w:t+T-w-1}$. Each decoder block then updates the hidden representation via the 2DMA layer
\begin{equation}
\boldsymbol{\phi}^i_t = \operatorname{2DMA}\left(\mathbf{h}_{t+w:t+T-w-1} \right).
\end{equation}

Then, the representation of each appliance is updated by a fully connected feed-forward network 
\begin{equation}
\mathbf{h}^i_t=\operatorname{LayerNorm}\left(\boldsymbol{\phi}^i_t + \max\left(0, \boldsymbol{\phi}^i_t\mathbf{W}^i_1 \right)\mathbf{W}_2^i\right),
\end{equation}
where we omit the bias term for brevity. 

In the last decoder block, we further split the network for each appliance in order to explicitly exploit power consumption and on/off state \cite{shin2019subtask}: 
\begin{subequations} 
\begin{align}
\mathbf{h}^i_{c,t}&=\operatorname{LayerNorm}\left(\boldsymbol{\phi}^i_t + \max\left(0, \boldsymbol{\phi}^i_t\mathbf{W}^i_{c,1} \right)\mathbf{W}_{c,2}^i\right),\\
\mathbf{h}^i_{r,t}&=\operatorname{LayerNorm}\left(\boldsymbol{\phi}^i_t + \max\left(0, \boldsymbol{\phi}^i_t\mathbf{W}^i_{r,1} \right)\mathbf{W}_{r,2}^i\right).
\end{align}
\end{subequations}

In the regression subnetwork, a fully connected layer with the rectified linear unit (ReLU) activation function is used to transform the hidden information to the estimated power consumption for $i$-th appliances at each time step:
\begin{equation} \label{eq:output}
    \hat{p}_t^i=\text{ReLU}(\mathbf{h}^i_{r,t}\mathbf{W}_{r}^i)\mathbf{V}_r^i.
\end{equation}

Similarly, in the classification subnetwork, a fully connected layer with the sigmoid activation function is used to transform the hidden information to the estimated on/off status for $i$-th appliances at each time step:
\begin{equation} \label{eq:output2}
    \hat{o}_t^i=\text{Sigmoid}(\mathbf{h}^i_{c,t}\mathbf{W}_c^i)\mathbf{V}_c^i.
\end{equation}

The final estimated power consumption for appliance $i$ is calculated by $\hat{y}_t^i = \hat{p}_t^i \times \hat{o}_t^i$.

\subsection{Loss Function}
In this work, we extend the loss function in \cite{shin2019subtask} to satisfy the MAT setup, which is defined given as:
\begin{equation} \label{eq:Loss}
    \mathcal{L} = \sum_{i=1}^{|\mathcal{K}|}(\mathcal{L}_{\text{output}}^i + \mathcal{L}_{\text{on}}^i),
\end{equation}
where $\mathcal{L}_{\text {output }}$ is the mean squared error (MSE) of the network final output:
\begin{equation}
\mathcal{L}_{\text {output }}^i=\frac{1}{T} \sum_t^T\left(y_t^i-\hat{p}_t^i \hat{o}_t^i\right)^2,
\end{equation}
and $\mathcal{L}_{\text {on }}$ is the binary cross entropy (BCE) of the classification subnetwork's result:
\begin{equation}
\mathcal{L}_{\text{on}}^i=-\frac{1}{T} \sum_{t=1}^T\left(o_t^i \log \hat{o}_t^i+\left(1-o_t^i\right) \log \left(1-\hat{o}_t^i\right)\right).
\end{equation}

\section{Experiment Setup}

\subsection{Data Preprocessing} \label{REDD}
The proposed algorithms are tested on two real-world datasets: Reference Energy Disaggregation Data Set (REDD) \cite{kolter_redd_2011} and UK-DALE \cite{kelly2015uk}. The REDD dataset comprises 6 US household aggregate load consumption data with time ranges varying from 23 to 48 days. The time resolution of aggregate signal is 1 second while 3 seconds for appliance-wise signals. To ensure the consistency of time alignment, the aggregate signal is downsampled to 3 seconds to match the appliance-wise sampling rate. On the other hand, the UK-DALE dataset includes 5 UK households, with house 1 having more than four years of recordings. The time resolution for aggregate signal and appliance-wise signals is 6 seconds.
The appliances of interest include fridge, dishwasher, microwave and washer dryer for REDD while an additional appliance kettle is being considered for UK-DALE. 
\\
\indent We follow the data preprocessing procedure in \cite{shin2019subtask}. In that study, each pair of the aggregate-appliance signal is processed independently, resulting in potentially different timestamps for the processed appliance sequences. However, for the multi-appliance-task framework, it is crucial to have a correct time alignment between each appliance signal and its corresponding aggregate one. To ensure that, we first utilize the merged dataset given by NILMTK \cite{batra2014nilmtk}, where the aggregated and individual appliance active power consumptions are merged into a single table based on the timestamps. Next, we split the table to exclude timestamps with 20 continuous missing values or 1200 continuously unchanged values for any signals. We then use the backward filling method to fill in the remaining missing values. We only consider sub-tables with more than one hour of duration to ensure a sufficient amount of data for further analysis. The entire dataset is normalized (divided by 612) in the same fashion as suggested by \cite{shin2019subtask}.

\subsection{Three Scenarios}
We consider three scenarios in this paper: Scenario one (S1) serves as a baseline when the entire dataset is available, following the common practice of prior works. For the REDD dataset under S1, houses 2-6 are used for training while house 1 is for testing; see also \cite{shin2019subtask} and \cite{chen_2020}. We select parts of data from houses 2 and 3 for validation because they have the active state for all the appliances to be estimated. Scenario two (S2) is designed to evaluate the proposed method with limited labeled data. We consider this scenario because obtaining long-term household-level labeled data is often challenging in practice. Specifically, we are confronted with an unfavorable scenario where only a single day's worth of labeled data is accessible for training, while another day's worth of data is utilized for validation and testing on an unseen household. Scenario three (S3) is built upon S2 by incorporating limited appliance activation measurements to augment the training samples. Table \ref{table:simulation_detail} presents an overview of the training, validation, and testing datasets used in this study.

\begin{table}[!tb]
\centering
\caption{Training, validation, and testing datasets for REDD and UK-DALE.}
\label{table:simulation_detail}
\renewcommand{\arraystretch}{1.2}
\resizebox{0.48\textwidth}{!}{%
\begin{tabular}{llcl}
\hline
                        &            & House \# & Time index                                \\ \hline
\multirow{3}{*}{REDD}   & Training   & 3        & 2011-04-21 19:41:24 - 2011-04-22 19:41:21 \\
                        & Validation & 3        & 2011-05-23 10:31:24 - 2011-05-24 10:31:21 \\
                        & Testing    & 1        & 2011-04-18 09:22:12 - 2011-05-23 09:21:51 \\ \hline
\multirow{3}{*}{UK-DALE} & Training   & 1        & 2017-04-23                                \\
                        & Validation & 1        & 2017-04-25                                \\
                        & Testing    & 2        & 2017-04-12 - 2017-04-25                   \\ \hline
\end{tabular}%
}
\end{table}

\subsection{Model Details}
The SGN-Conv model is implemented based on the specifications in \cite{shin2019subtask}. In addition, we also explore a variant of SGN called SGN-LSTM, where each subnetwork is composed of five layers of BiLSTM followed by two linear layers with a hidden size of 32. The MAT-Conv model uses the convolutional layers of SGN-Conv as its encoder, while the LSTM layers in SGN-LSTM serve as the encoder for MAT-LSTM. For each target appliance, three decoder blocks are employed in this study.
\\
\indent For REDD, the length of the input sequence is fixed at 864 while the output length is 64. This results in an additional window size $w=400$. For UK-DALE, the input and output lengths are 464 and 64, respectively. Early stopping is used as a training criteria with a patience of 30. This means that the training procedure stops if the validation performance does not improve for 30 epochs. The detailed training algorithm with our sample augmentation and early stopping is shown in Algorithm \ref{alg:training}. We use the MSE loss for the regression subnetwork and the BCE loss for the classification subnetwork. Based on PyTorch 1.6.0, all models are trained via the Adam optimizer with an initial learning rate of 0.001. 
% The models are implemented using PyTorch 1.6.0, and the code is publicly available at [placeholder].

\begin{algorithm}[!bt]
\caption{Training Algorithm with Sample Augmentation (SA) and Early Stopping}
\label{alg:training}
\SetKwInOut{Input}{Input}
\SetKwInOut{Output}{Output}

\Input{
Training set $\mathcal{D}$, validation set $\mathcal{D}_{\mathrm{val}}$, number of epochs $E$, patience parameter $p$, model $f_{\theta}$, SA algorithm $\mathrm{Algo}_{\mathrm{SA}}$, SA flag $f_{\mathrm{SA}}$}
\Output{Trained model $f_{\theta}$}

Initialize model parameters\;
Set early stopping counter $c=0$\;
Assign a large enough number to the best loss $L_{\mathrm{best}}$\; 
\For{$e \in {1,2,...,E}$}{
    \For{each batch $\Bar{\mathcal{D}} \in \mathcal{D}$}{
        \If{$f_{\mathrm{SA}}=\texttt{true}$}{
            $\Bar{\mathcal{D}} \gets \mathrm{Algo}_{\mathrm{SA}}(\Bar{\mathcal{D}})$; \tcp{Algorithm \ref{alg:dataAugAl}}
            }
        $\hat{y}  \gets f_{\theta}(\Bar{\mathcal{D}})$\;
        Compute training loss $L_{\mathrm{train}}$\;
        Compute the gradient of loss w.r.t. $\theta$\;
        Update parameters in $f_{\theta}$\;
        }
    Compute validation loss $L_{\mathrm{val}}$ on $\mathcal{D}_{\mathrm{val}}$\;
    \If{$L_{\mathrm{val}}$ $<$ $L_{\mathrm{best}}$}{
        Save model $f_{\theta}$\;
        Update $L_{\mathrm{best}}$ = $L_{\mathrm{val}}$\;
        Reset $c$ = 0\;
    }
    \Else{
        $c$ = $c$ + 1\;
        \If{$c$ $\geq$ $p$}{
            Stop training\;
        }
    }
}
Return the saved model $f_{\theta}$.
\end{algorithm}

\subsection{Performance Metrics}

Mean absolute error (MAE) and signal aggregate error (SAE) are used as evaluation metrics of disaggregation performance for each appliance. They are defined as follows \cite{shin2019subtask}:
\begin{subequations} \label{eq:evaluation}
\begin{align}
    \mathrm{MAE} &= \frac{1}{H}\sum_{t=1}^H \left|y_t - \hat{y}_t\right|\\
    \mathrm{SAE} &= \frac{1}{S} \sum_{\tau=0}^{S-1}\frac{1}{M}\left| y_{\tau} - \hat{y}_{\tau}\right|,
\end{align}
\end{subequations}
where $y_t$ and $\hat{y}_t$ are the ground truth and estimated power consumption at time $t$, respectively. $H$ is the length of the test horizon. For SAE, the test horizon is split into $S$ disjoint sub-horizons. The length of each sub-horizon is $M$. $y_{\tau} := \sum_{t=1}^{M} y_{M\tau+t}$ is the total power consumption across $M$ timestamps in the $\tau$-th sub-horizon and $\hat{y}_{\tau}$ is the corresponding estimated value. In this study, for both datasets we set $M=1200$ and $S=	\lfloor\frac{H}{M}\rfloor$ (i.e., the integer part of $\frac{H}{M}$). 
% as being used in \cite{shin2019subtask} and \cite{chen_2020}.

The classification performance is evaluated by the $F_1$ score, which is defined as the harmonic mean of 
precision and recall:
\begin{equation} \label{eq:f1}
    F_1 = 2 \times \frac{\text { precision } \times \text { recall }}{\text { precision }+\text { recall }}.
\end{equation}
Precision is the ratio of true positives to the total number of positive predictions. Recall is the ratio of true positives to the total number of actual positive instances. Due to the inherent tradeoff between these two metrics, the F1 score turns out to be especially valuable when dealing with imbalanced datasets.

\section{Experiment Results}
In this section, we showcase the outcomes of three case studies, where the results have been averaged over three independent trials. These case studies demonstrate the effectiveness and merits of our proposed framework. Case 1 evaluates the performance of existing models on a limited labeled data scenario and compares them with our proposed solution that utilizes SA and MATNilm. Case 2 is an ablation study of the proposed approach. Moreover, we compare our SA approach with two state-of-the-art data augmentation methods on four different models: SGN, LDwA, VMA, and MATNilm. Case 3 uses the SGN model to show training performance on a full dataset versus a limited dataset with SA.

\subsection{Case 1: Performance Comparisons with Limited Training Data}
\begin{figure}[!tb]
\centering
\includegraphics[width=\columnwidth]{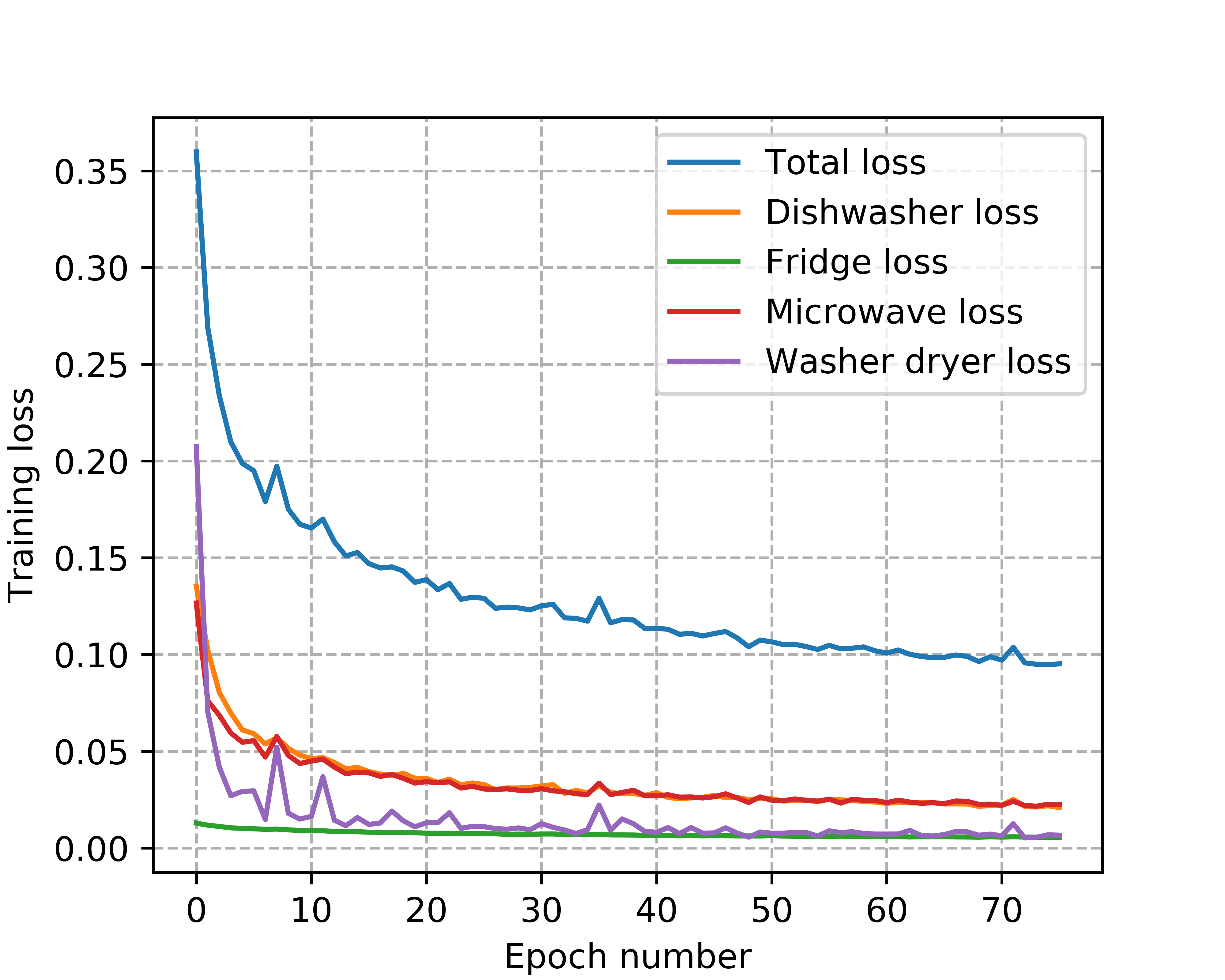}
\caption{The convergence of MAT-conv training losses.}
\label{fig:loss}
\end{figure}

In this test case, we focus on S2 and S3 where only limited training data are available. S2 directly applies a learning model to a limited labeled data scenario while S3 is our proposed solution: MATNilm with SA. Regarding the learning models, both SGN and MATNilm can be implemented with a convolutional or LSTM shared layer. Table \ref{table:c1} presents the comparison of those four model variants. The performance metrics include MAE, SAE and F1 scores. The average improvement for MAE is defined as
\begin{equation}
\mathrm{Imp} = \frac{\mathrm{MAE(SGN)} - \mathrm{MAE(MAT)}}{\mathrm{MAE(SGN)}} \times 100\%\, .   
\end{equation}
Similarly, we can calculate the average improvements for SAE and F1 scores.
\\
\indent First, our proposed solution MAT-Conv and MAT-LSTM consistently outperform the SGN counterparts in terms of MAE (SAE) scores. For REDD, MAT-Conv and MAT-LSTM models with SA significantly reduce MAE, with an average improvement of 52.23\% (58.84\%) and 56.04\% (63.05\%), respectively. We observe similar performance gains for UK-DALE. For microwave in UK-DALE, the MAE values for different models are at the same level. However, the performance of F1 and SAE is much better for S3. 
\\
\indent It is worth noting that there are instances of divergent performance between the MAE and SAE metrics. For example, MAT-Conv has a higher MAE than MAT-LSTM for fridge in REDD, while the SAE performance is the opposite. In such cases, the model having lower SAE is better at accurately disaggregating the appliance's power consumption over time, but is not necessarily very effective in estimating the power at each timestamp. 
% Specifically, MAT-Conv appears to suffer from short-term but significant errors more frequently than MAT-LSTM, but nevertheless achieves a better overall estimate of the energy consumed over time.
\\
\indent Second, S3 outperforms S2 in F1 score for both datasets. Recall that S3 has more training samples augmented with appliance activation profiles while S2 lacks representative examples of appliance's power consumption patterns. The limited data scenario makes it harder for the learning models to differentiate between different appliance statuses. This in turn validates the strong capability of our solution to correctly detect on/off status that leads to more accurate disaggregation results.
In a nutshell, the results demonstrate that our proposed approach of using MATNilm with SA can substantially improve disaggregation performance, particularly when training data are limited.
\\
\indent Finally, Fig.~\ref{fig:loss} shows the convergence of the MAT-conv model. It can be seen that the total loss and all individual appliance-level losses converge in about 80 epochs.

\begin{table*}[!tb]
\centering
\caption{REDD and UK-DALE: Performance comparisons of existing models and the proposed framework with limited data. \\
DW, FG, MW, KT, WD are the acronyms for ``Dishwasher", ``Fridge", ``Microwave", ``Kettle", ``Washer Dryer".\\
Ave and Imp stand for ``Average score" and ``Average improvement".}
\label{table:c1}
\renewcommand{\arraystretch}{1.2}
\resizebox{\textwidth}{!}{%
\begin{tabular}{clcccccccccccccc}
\hline
\multirow{2}{*}{Metric} &
  \multirow{2}{*}{Model} &
  \multirow{2}{*}{Scenario} &
  \multicolumn{6}{l}{REDD} &
  \multicolumn{7}{l}{UK-DALE} \\ \cline{4-16} 
 &
   &
   &
  DW &
  FD &
  MW &
  WD &
  Ave &
  Imp &
  DW &
  FD &
  MW &
  KT &
  WD &
  Ave &
  Imp \\ \hline
\multirow{4}{*}{MAE} &
  SGN - Conv \cite{shin2019subtask} &
  S2 &
  22.14 &
  39.01 &
  19.40 &
  40.13 &
  30.17 &
  - &
  19.81 &
  27.05 &
  8.03 &
  15.09 &
  20.92 &
  18.18 &
  - \\
 &
  SGN - LSTM &
  S2 &
  21.98 &
  41.46 &
  21.29 &
  43.91 &
  32.16 &
  - &
  21.40 &
  32.35 &
  7.90 &
  7.22 &
  21.24 &
  18.02 &
  - \\
 &
  MAT - Conv &
  S3 &
  \textbf{8.44} &
  19.38 &
  13.40 &
  \textbf{16.44} &
  14.41 &
  52.23\% &
  10.88 &
  17.06 &
  \textbf{7.08} &
  5.95 &
  6.52 &
  9.50 &
  47.75\% \\
 &
  MAT - LSTM &
  S3 &
  9.12 &
  \textbf{17.86} &
  \textbf{12.49} &
  17.08 &
  \textbf{14.14} &
  56.04\% &
  \textbf{6.51} &
  \textbf{15.86} &
  8.20 &
  \textbf{5.36} &
  \textbf{5.37} &
  \textbf{8.26} &
  54.17\% \\ \hline
\multirow{4}{*}{SAE} &
  SGN - Conv \cite{shin2019subtask} &
  S2 &
  21.83 &
  25.78 &
  17.87 &
  39.11 &
  26.15 &
  - &
  15.00 &
  13.07 &
  8.06 &
  13.09 &
  20.79 &
  14.00 &
  - \\
 &
  SGN - LSTM &
  S2 &
  21.61 &
  31.71 &
  17.89 &
  39.49 &
  27.67 &
  - &
  19.52 &
  16.23 &
  7.49 &
  5.49 &
  20.24 &
  13.79 &
  - \\
 &
  MAT - Conv &
  S3 &
  \textbf{6.94} &
  \textbf{12.30} &
  10.47 &
  \textbf{13.35} &
  10.76 &
  58.84\% &
  9.44 &
  6.93 &
  \textbf{5.53} &
  \textbf{3.01} &
  4.54 &
  5.89 &
  57.92\% \\
 &
  MAT - LSTM &
  S3 &
  8.20 &
  12.67 &
  \textbf{9.81} &
  14.29 &
  \textbf{10.23} &
  63.05\% &
  \textbf{5.39} &
  \textbf{6.79} &
  6.70 &
  3.34 &
  \textbf{3.65} &
  \textbf{5.18} &
  62.48\% \\ \hline
\multirow{4}{*}{F1} &
  SGN - Conv \cite{shin2019subtask} &
  S2 &
  0.19 &
  0.80 &
  0.10 &
  0.33 &
  0.36 &
  - &
  0.85 &
  0.65 &
  0.00 &
  0.39 &
  0.08 &
  0.39 &
  - \\
 &
  SGN - LSTM &
  S2 &
  0.23 &
  0.71 &
  0.08 &
  0.61 &
  0.41 &
  - &
  0.76 &
  0.60 &
  0.18 &
  0.85 &
  0.09 &
  0.50 &
  - \\
 &
  MAT - Conv &
  S3 &
  0.80 &
  0.88 &
  0.66 &
  \textbf{0.67} &
  0.75 &
  110.98\% &
  0.82 &
  0.83 &
  \textbf{0.67} &
  \textbf{0.91} &
  0.79 &
  0.80 &
  104.41\% \\
 &
  MAT - LSTM &
  S3 &
  \textbf{0.82} &
  \textbf{0.91} &
  \textbf{0.74} &
  0.64 &
  \textbf{0.78} &
  89.23\% &
  \textbf{0.91} &
  \textbf{0.84} &
  0.65 &
  0.90 &
  \textbf{0.82} &
  \textbf{0.83} &
  66.76\% \\ \hline
\end{tabular}%
}
\end{table*}

\subsection{Case 2: Ablation Study and DA comparisons}
Table \ref{table:ablation} summarizes the MAE scores obtained by the ablation study for the sample augmentation and MATNilm with convolutional layers as the encoder. The study evaluates the performance of the framework with different combinations of SGN, sample augmentation (SA), multi-appliance task (MT), temporal attention (TA), and appliance attention (AA). The results show that SA significantly reduces the MAE values. The errors further decrease by incorporating the temporal and appliance attention to the vanilla MT network. This validates the effectiveness of each proposed module in our framework.
\\
Using four models, namely SGN \cite{shin2019subtask}, LDwA \cite{piccialli2021improving}, VMA \cite{schirmer2021binary}, and MATNilm, we conducted further comparisons between our sample augmentation approach and two state-of-the-art NILM data augmentation methods \cite{kong2019practical} and \cite{rafiq2021generalizability}. However, since both of these competing alternatives were not originally designed to address the limited data scenario or the multi-appliance approach, certain modifications were necessary to tailor them to our specific problem setting. The original data augmentation methods were designed to augment a single target appliance and generated (aggregate, appliance) pairs of signals for each appliance. This resulted in misalignment among the aggregate signals and rendered them unsuitable for the multi-appliance approach, where the aggregate signal remains the same for each appliance. To ensure a fair comparison, we employed an aligned aggregate signal dataset for both the single-appliance and multi-appliance models. To adapt Kong's method \cite{kong2019practical}, we duplicated the one-day labeled data 20 times and applied the same augmentation steps to all target appliances, following the reference algorithms in \cite{kong2019practical}. For \cite{rafiq2021generalizability}, we first duplicated the one-day labeled data three times. Then, we generated an augmented synthetic dataset for each appliance and concatenated them to create the final dataset for training and evaluation. These modifications allowed us to compare the performance of our sample augmentation approach against these state-of-the-art methods in a fair and consistent manner.
\\
\indent Table \ref{table:case22} summarizes the MAE performance for the aforementioned comparisons. Notably, our proposed MATNilm with SA achieves the lowest MAE scores for all appliances, with an average MAE of 14.41. When considering a single model, our proposed SA demonstrates significant performance improvement over the other two competing data augmentation (DA) methods \cite{kong2019practical, rafiq2021generalizability}, except for VMA. This suggests that for deep models, SA is capable of generating richer data that benefits the model's learning process. However, in the case of VMA, which has a shallow model structure, its learning ability is limited, and therefore it does not derive significant benefits from the augmented training data.
\\
\indent Without the two-dimensional attention, MAT and VMA have inferior performance compared with the single-appliance SGN. A possible reason is that the shared and split structures of the two multi-appliance approaches cannot adequately capture the complex cross-appliance information, which is necessary for yielding  better disaggregation outcomes. Furthermore, employing a single model for all appliances does not offer much flexibility when it comes to fine-tuning the best model for each appliance.
\\
\indent To sum up, the numerical results indicate that SA and MAT with 2DMA are crucial components for achieving accurate appliance-level energy disaggregation. The ablation study provides us an insight into the relative importance of different modules of the proposed framework.

\begin{table}[!tb]
\centering
\caption{REDD: MAE results for ablation study of the proposed framework. MT, TA, AA are the acronyms for ``Multi-appliance-task", ``Temporal attention", ``Appliance attention".}
\label{table:ablation}
\renewcommand{\arraystretch}{1.2}
% \resizebox{\textwidth}{!}{%
\begin{tabular}{ccccccccc}
\hline
SGN                       & SA                        & MT & TA & AA & DW & FG & MW & WD \\ \hline
\checkmark &                           &    &    &    & 22.14      & 39.01  & 19.40     & 40.13  \\
\checkmark & \checkmark &    &    &    & 14.31      & 28.68  & 16.90     & 21.79  \\
\checkmark & \checkmark & \checkmark &                           &                           & 24.26 & 42.83 & 19.78 & 34.18 \\
\checkmark & \checkmark & \checkmark & \checkmark &                           & 9.73  & 20.08 & \textbf{13.33} & 20.12 \\
\checkmark & \checkmark & \checkmark & \checkmark & \checkmark & \textbf{8.44}  & \textbf{19.38} & 13.40 & \textbf{16.44} \\ \hline
\end{tabular}%
% }
\end{table}

\begin{table}[!tb]
\centering
\renewcommand{\arraystretch}{1.2}
\caption{REDD: MAE results for performance comparison between the proposed SA and two existing augmentation methods. }
\label{table:case22}
% \resizebox{\textwidth}{!}{%
\begin{tabular}{ccccccc}
\hline
DA                   & Model   & DW & FD & MW & WD & Average \\ \hline
\multirow{4}{*}{Kong\cite{kong2019practical}} & SGN\cite{shin2019subtask}  &22.14	&39.01	&19.40	&40.13	&30.17\\
                    & LDwA\cite{piccialli2021improving}     & 17.62	&39.44	&18.93	&39.21	&28.80  \\
                     & VMA\cite{schirmer2021binary}     & 26.53      & 42.14  & 29.85     & 49.96  & 37.12   \\
                     & MATNilm & 11.72      & 32.61  & 26.93     & 46.66  & 29.48   \\ \hline
\multirow{4}{*}{Rafiq\cite{rafiq2021generalizability}} & SGN\cite{shin2019subtask}     & 21.30	&42.81	&19.77	&33.29	&29.29   \\
                    & LDwA\cite{piccialli2021improving}     & 22.14	&39.01	&19.40	&40.13	&30.17 \\
                     & VMA\cite{schirmer2021binary}     & 24.59      & 49.62  & 18.66     & 41.59  & 33.61   \\
                     & MATNilm & 24.77      & 44.68  & 20.12     & 42.31  & 32.97   \\ \hline
\multirow{4}{*}{SA}  & SGN\cite{shin2019subtask}     & 14.31      & 28.68  & 16.90     & 21.79  & 20.42   \\
                    & LDwA\cite{piccialli2021improving}     &16.41	&28.32	&19.34	&27.63	&22.93  \\
                     & VMA\cite{schirmer2021binary}     & 25.41      & 47.69  & 18.47     & 45.37  & 34.24   \\
                     & MATNilm & \textbf{8.44}       & \textbf{19.38}  & \textbf{13.40}     & \textbf{16.44}  & \textbf{14.41}   \\ \hline
\end{tabular}%
% }
\end{table}

% Please add the following required packages to your document preamble:
% \usepackage{multirow}
% \usepackage{graphicx}

% Of course, this relatively naive approach to synthesising aggregate data ignores a lot of structure that appears in real aggregate data. For example, the kettle and toaster might often appear within a few minutes of each other in real data, but our simple ‘simulator’ is completely unaware of this sort of structure.

% use static variables to infer this in the future.

% \ref{fig:comparison}. 

\subsection{Case 3: Comparison with Training on Full Dataset}

In practice, obtaining household labeled data is often challenging. Appliance operation profiles are comparatively easier to get. Hence, we aim to investigate whether the model can achieve comparable performance with limited labeled data and appliance operation profiles, compared with the case of several weeks of data from multiple houses. Specifically, we  compare the results of models trained in S1 and S3. The latter uses only one day of labeled data and appliance operation profiles, which are extracted from the same training houses as in S1, for sample augmentation.
\\
\indent Table \ref{table:case3} presents the performance comparison in scenarios S1 and S3 for the REDD dataset. The results indicate that limited data with SA can significantly improve the F1 score for most appliances, except for the fridge. This improvement can be attributed to the fact that SA tends to balance the on/off status for each appliance, which also results in better performance in terms of MAE and SAE. However, since the fridge operates quite uniformly over time, there is no significant improvement in its performance with SA.

\begin{table}[!tb]
\centering
\renewcommand{\arraystretch}{1.2}
\caption{REDD: Comparison of the SGN model trained on the full dataset (S1) versus on the limited dataset with SA (S3).}
\label{table:case3}
% \resizebox{\textwidth}{!}{%
\begin{tabular}{ccccccc}
\hline
Metric               & Scenario & DW    & FD    & MW    & WD    & Average \\ \hline
\multirow{2}{*}{MAE} & S1       & 19.51 & \textbf{27.27} & 23.61 & 39.22 & 27.40   \\
                     & S3       & \textbf{14.31} & 28.68 & \textbf{16.90} & \textbf{21.79} & \textbf{20.42}   \\ \hline
\multirow{2}{*}{SAE} & S1       & 18.92 & 17.23 & 18.33 & 34.79 & 22.32   \\
                     & S3       & \textbf{12.73} & \textbf{16.77} & \textbf{11.94} & \textbf{16.12} & \textbf{14.39}   \\ \hline
\multirow{2}{*}{F1}  & S1       & 0.34  & 0.83  & 0.04  & 0.29  & 0.38    \\
                     & S3       & \textbf{0.70}  & \textbf{0.85}  & \textbf{0.65}  & \textbf{0.59}  & \textbf{0.70}    \\ \hline
\end{tabular}%
% }
\end{table}

\section{Conclusion}

In this paper, we propose a solution for the NILM problem with limited training data. Specifically, the multi-appliance-task framework adopts a shared-hierarchical split structure for each appliance based on its power consumption (regression) and status (classification) tasks. A two-dimensional attention mechanism is developed to capture power consumption relationships across each appliance and time step. Instead of generating a time-continuously synthetic dataset,  we design a dynamic sample augmentation algorithm, where augmented training samples are randomly generated for each appliance in each mini-batch. Experiment results show a significant performance boost by the proposed solution. Moreover, the simulation results also reveal the importance of collecting individual appliance operation profiles, which may open up a new research direction for the NILM.

\bibliographystyle{IEEEtran}
\bibliography{ref,IEEEabrv}
% that's all folks
\end{document}